\documentclass[letterpaper, 10 pt, conference]{ieeeconf}
\IEEEoverridecommandlockouts                              
\usepackage{mathrsfs}
\usepackage{amsfonts}
\usepackage{amsmath,amssymb}
\usepackage{graphicx}
\usepackage{subfigure}
\usepackage{epstopdf}
\usepackage{subfigure}
\usepackage{epic}
\usepackage{tikz}
\usepackage{bm}
\usepackage{algorithm}
\usepackage{algpseudocode}
\usepackage{graphics} 
\usepackage{color}
\usepackage[justification=centering]{caption}
\usetikzlibrary{arrows,shapes,chains}
\usepackage{cite}
\usepackage{algorithm}
\usepackage{algpseudocode}

\floatname{algorithm}{Algorithm}

\newtheorem{definition}{Definition}

\newtheorem{remark}{Remark}

\newtheorem{proposition}{Proposition}

\begin{document}
%
\title{\LARGE{\bf{A fully distributed motion coordination strategy for multi-robot systems with local information}}}
%
%

\author{Pian Yu and Dimos V. Dimarogonas
\thanks{This work was supported in part by the Swedish Research Council (VR), the Swedish Foundation for Strategic Research (SSF) and the Knut and Alice Wallenberg Foundation (KAW).}
\thanks{The authors are with School of Electrical Engineering and Computer Science, KTH Royal Institute of Technology, 10044 Stockholm, Sweden.
        {\tt\small piany@kth.se, dimos@kth.se}}
}

\maketitle

\begin{abstract}
This paper investigates the online motion coordination problem for a group of mobile robots moving in a shared workspace. Based on the realistic assumptions that each robot is subject to both velocity and input constraints and can have only local view and local information, a fully distributed multi-robot motion coordination strategy is proposed. Building on top of a cell decomposition, a conflict detection algorithm is presented first. Then, a rule is proposed to assign dynamically a planning order to each pair of neighboring robots, which is deadlock-free. Finally, a two-step motion planning process that combines fixed-path planning and trajectory planning is designed. The effectiveness of the resulting solution is verified by a simulation example.
\end{abstract}
%

\IEEEpeerreviewmaketitle

\section{Introduction}

One challenge for multi-robot systems (MRSs) is the design of coordination strategies between robots that enable them to perform operations safely and efficiently in a shared workspace while achieving individual/group motion objectives \cite{Yan13}. This problem was originated from 1980s and has been extensively investigated since. In recent years, the attention that has been put on this problem has grown significantly due to the emergence of new applications, such as smart transportation and service robotics. The existing literature can be divided into two categories: \emph{path coordination} and \emph{motion coordination}. The former category plans and coordinates the entire paths of all the robots in advance, while the latter category focuses on decentralized approaches that allow robots to resolve conflicts online as the situation occurs\footnote{In some literatures, these two terms are also used interchangeably. In this paper, we try to distinguish between the two as explained above.} \cite{Parker09}. This paper aims at developing a fully distributed strategy for  multi-robot motion coordination (MRMC).

Depending on how the controller is synthesized for each robot, the literature concerning MRMC can further be classified into two types: the reactive approach and the planner-based approach. Typical methods that generate reactive controllers consist of potential-field approach \cite{Khatib1986}, sliding mode control \cite{Gracia2013} and control barrier functions \cite{Wang17}. These reactive-style methods are fast and operate well in real-time. However, it is well-known that these methods are sensitive to deadlocks that are caused by local minima. Moreover, guidance for setting control parameters is not analyzed formally when explicit constraints on the system states and/or inputs are presented \cite{Parker09}. Apart from the above, other reactive methods include the generalized roundabout policy \cite{Pallottino07} and a family of biologically inspired methods \cite{Bekey05}.

An early example of the planner-based method is the work of Azarm and Schmidt \cite{Azarm97}, where a framework for online coordination of multiple mobile robots was proposed. Based on this framework, various applications and different motion planning algorithms are investigated. Roughly speaking, the motion planning algorithms used in planner-based approaches can be divided into two types: fixed-path planning \cite{Simon02, Liu17} and trajectory planning \cite{Azarm97,Guo02,Stentz97}, while the former one differs from the latter one in that the motions for individual robots are fixed along specific paths. Guo and Parker \cite{Guo02} proposed a MRMC strategy based on the D$^*$ algorithm \cite{Stentz97}. In this work, each robot has an independent goal position to reach and know all path information. In \cite{Sheng06}, a distributed bidding algorithm was designed to coordinate the movement of multiple robots, which focuses on area exploration. In the work of Liu \cite{Liu17}, conflict resolution at intersections was considered for connected autonomous vehicles, where each vehicle is required to move along a pre-planned path. In general, the fixed-path planning method is more efficient. Its major disadvantage, however, lies in the fact that it fails more often. A literature review on MRMC can be found in \cite{Yan13}.

In this paper, we investigate the MRMC problem on a realistic setup. Robots are assumed to have limited sensing capabilities and both velocity and input constraints are considered. Conflicts are assumed to be local and can occur at arbitrary locations in the workspace. To cope with this setup, a fully distributed MRMC strategy is proposed. The contributions of this paper can be summarized as follows. Building on top of a cell decomposition, a formal definition of spatial-temporal conflict is introduced first. This definition characterizes when replanning is required for each robot. Then, a simple rule is proposed for online planning order assignment, which is deadlock-free when only local information is available to each robot. Finally, a two-step motion planning process is proposed, i.e., the fixed-path planning is activated first while the trajectory planning is activated if and only if the fixed-path planning returns no feasible solution, which allows us to leverage both the benefits of fixed-path planning and trajectory planning.

The remainder of the paper is organized as follows. In Section
II, notation and preliminaries on graph theory are introduced. Section
III formalizes the considered problem. Section IV presents
the proposed solution in detail, which is verified by
simulations in Section V. Conclusions are given in Section
VI.

\section{Preliminaries}

\subsection{Notation}
Let $\mathbb{R}:=(-\infty, \infty)$, $\mathbb{R}_{\ge 0}:=[0, \infty)$, and $\mathbb{Z}_{\ge 0}:=\{0,1,2,\ldots\}$. Denote $\mathbb{R}^n$ as the $n$ dimensional real vector space, $\mathbb{R}^{n\times m}$ as the $n\times m$ real matrix space.  Let $\left|\lambda\right|$ be the absolute value of a real number $\lambda$, $\|x\|$ and $\|A\|$ be the Euclidean norm of vector $x$ and matrix $A$, respectively. Given a set $\Omega$, $2^{\Omega}$ denotes its powerset and $|\Omega|$ denotes its cardinality. Given two sets $\Omega_1, \Omega_2$, the set $\mathcal{F}(\Omega_1, \Omega_2)$ denotes the set of all functions from $\Omega_1$ to $\Omega_2$. The operators $\cup$ and $\cap$ represent set union and set intersection, respectively. In addition, we use $\wedge$ to denote the logical operator AND and $\vee$ to denote the logical operator OR. The set difference $A\setminus B$ is defined by $A\setminus B:=\{x: x\in A \;\wedge\; x\notin B\}$.
Given a point $x\in \mathbb{R}^n$ and a constant $r\ge 0$, the notation $\mathcal{B}(x, r)$ represents a ball area centered at point $x$ and with radius $r$.

\subsection{Graph Theory}

Let $\mathcal{G}=\{\mathcal {V}, \mathcal {E}\}$ be a digraph with the set of nodes $\mathcal {V}={1, 2,\dots, N}$, and $\mathcal{E}\subseteq \{(i,j): i,j\in \mathcal {V},j \ne i\}$ the set of edges. If $(i,j)\in \mathcal {E}$, then node $j$ is called a neighbor of node $i$ and node $j$ can receive information from node $i$. The neighboring set of node $i$ is denoted by $\mathcal{N}_i = \{j \in \mathcal {V} | (j, i)\in \mathcal {E}\}$. A graph is called undirected if $(i,j)\in \mathcal {E} \Leftrightarrow (j,i)\in \mathcal {E}$, and a
graph is called connected if for every pair of nodes $(i, j)$, there exists a path which connects $i$ and $j$, where a path is an ordered list of edges such that the head of each edge is equal to the tail of the following edge.

\section{Problem Formulation}\label{section_pf}

Consider a group of $N$ mobile robots, whose dynamics are given by:
\begin{equation}\label{x}
\begin{array}{l}
\dot{x}_i(t) = v_i(t)\cos(\theta_i(t)),\\
\dot{y}_i(t) = v_i(t)\sin(\theta_i(t)),\\
\dot{\theta}_i(t) = \omega_i(t),\\
\dot{v}_i(t) = F_i(t),\\
\dot{\omega}_i(t) = \tau_i(t),
 \quad i = 1,2, \ldots, N.
\end{array}
\end{equation}
where $p_i:=(x_i, y_i)$ is the Cartesian position, $\theta_i$ is the orientation, $v_i, \omega_i$ are respectively the tangential velocity and the angular velocity, $F_i$ is the force input, and $\tau_i$ is the torque input of robot $i$. For convenience, we define $\xi_i:=(x_i, y_i, \theta_i, v_i, \omega_i)$ and $u_i:=(F_i, \tau_i)$. Then, (\ref{x}) can be written as $\dot{\xi_i}=f(\xi_i, u_i)$, where $f(\xi_i, u_i)=(v_i\cos(\theta_i), v_i\sin(\theta_i), \omega_i, 0, 0)+(0, 0, 0, u_i)$. The velocity and input of each robot $i$ are subject to the constraints
\begin{equation}\label{cons}
  \begin{aligned}
  |v_i(t)|&\le v_i^{\max}, |\omega_i(t)|\le \omega_i^{\max}, \\ |F_i(t)|&\le F_i^{\max}, |\tau_i(t)|\le \tau_i^{\max}, \forall t\ge 0.
  \end{aligned}
\end{equation}
Let $\mathbb{V}_i:=\{v_i: |v_i|\le v_i^{\max}\}, \mathbb{W}_i:=\{\omega_i: |\omega_i|\le \omega_i^{\max}\}$ and $\mathbb{U}_i:=\{(F_i, \tau_i): |F_i|\le F_i^{\max}, |\tau_i|\le \tau_i^{\max}\}$.

Given a vector $\xi_i$, define the projection operator $\verb"proj"_{p_i}(\xi_i): \mathbb{R}^{5} \to \mathbb{R}^{2}$ as a mapping from $\xi_i$ to its first 2 components $p_i$. A curve ${\bm{\xi}}_i: [0, T[\to \mathbb{R}^{5}$ is said to be a trajectory of robot $i$ if there exists input $u_i(t)\in \mathbb{U}_i$ satisfying $\dot{\bm{\xi}}_i(t)=f({\bm{\xi}}_i(t), u_i(t))$ for all $t\in [0, T[$. A curve $\textbf{p}_i: [0, T[\to \mathbb{R}^{2}$ is a position trajectory of robot $i$ if $\textbf{p}_i(t)=\verb"proj"_{p_i}({\bm{\xi}}_i(t)), \forall t\in [0, T[$. Given a time interval $[t_1, t_2], t_1<t_2$, the corresponding position trajectory is denoted by $\textbf{p}_i([t_1, t_2])$.

Supposing that the sensing radius of each robot is the same, given by $R >0$, then the communication graph formed by the group of robots is undirected. The neighboring set of robot $i$ at time $t$ is given by $\mathcal{N}_i(t)=\{j: \|x_i(t)-x_j(t)\|\le R, j\in \mathcal{V}, j\neq i\}$, so that $j\in \mathcal{N}_i(t)\Leftrightarrow i\in \mathcal{N}_j(t), i\neq j, \forall t$. The group of robots are working in a common workspace $\mathbb{X}\subset \mathbb{R}^{2}$, which is populated with $m$ closed sets $O_i$, corresponding to obstacles. Let $\mathbb{O}={\cup}_{i} O_i$, then the free space $\mathbb{F}$ is defined as $\mathbb{F}:=\mathbb{X}\setminus \mathbb{O}$.

Each robot $i$ is subject to its own task specification $\varphi_i$ (in this work, we consider $\varphi_i$ to be a reach-avoid type of task expressed as a linear temporal logic (Chapter 5 \cite{Baier08}) formula). Given a position trajectory $\textbf{p}_i$, the satisfaction relation is denoted by $\textbf{p}_i\models \varphi_i$. Given the position $p_i$ of robot $i$, we refer to its \emph{footprint} $\phi(p_i)$ as the set of points in $\mathbb{X}$ that are occupied by robot $i$ in this position. The objective of the system is to ensure that the task specification $\varphi_i$ of each robot is satisfied efficiently (in the sense that a pre-defined objective function, e.g., $J_i$, is minimized), while safety (no inter-robot collision) of the system is guaranteed.

Note that in this paper, it is assumed that each robot is not aware of the existence of other robots. Moreover, each robot has only local view and local information. Under these settings, the MRMC problem has to be broken into local distributed motion coordination problems and solved online for individual robots. Let $\textbf{p}_j([t, t_j^*(t)]), j\in \mathcal{N}_i(t)$ be the local position trajectory of robot $j$ that is available to robot $i$ at time $t$, where $t_j^*(t)$ is determined by $t$. Then, the (online) motion coordination problem for robot $i$ is formulated as
\begin{subequations}\label{objec2}
\begin{eqnarray}
&&\hspace{-1cm}\min \quad  J_i(\xi_i, u_i)\\
&&\hspace{-1.2cm}\text{subject to}\nonumber \\
&&\hspace{-1cm} (\ref{x}), (\ref{cons}) \; \text{and} \; \textbf{p}_i\models \varphi_i, \label{c3}\\
&&\hspace{-1cm} \phi(p_i(t'))\cap \phi(p_j(t'))=\emptyset, \forall j\in \mathcal{N}_i(t), \forall t'\in [t, t_j^*(t)], \label{c2}
\end{eqnarray}
\end{subequations}
where constraint (\ref{c2}) means that two robots can not arrive at the same cell at the same time for all $t'\in [t, t_j^*(t)]$, thus guarantees no inter-robot collision occurs.

\section{Solution}

The proposed solution to the motion coordination problem (\ref{objec2}) consists of two layers: 1) an initialization layer and 2) an online coordination layer.

\subsection{Structure of each robot}

Before moving on, the structure of each robot is presented. Each robot $i$ is equipped with four modules, the decision making module, the motion planning module, the control module and the communication module. The first three modules work sequentially while the communication module works in parallel with the first three.

\begin{figure}
  \centering
  \includegraphics[width=.35\textwidth]{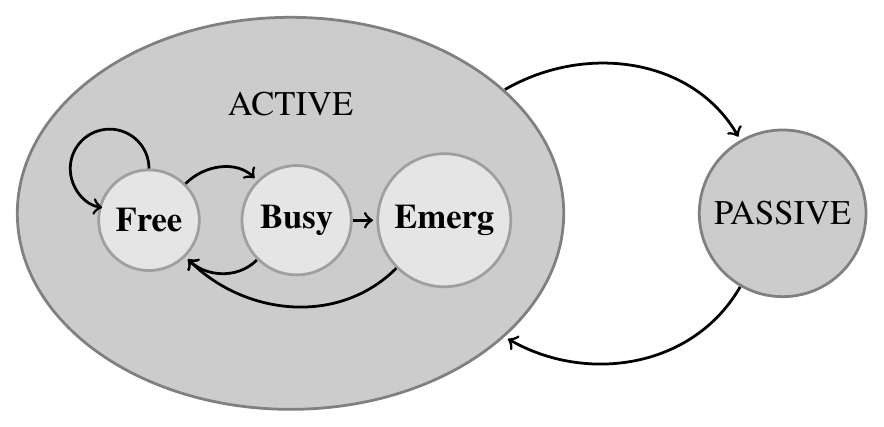}
  \caption{Transitions of robot $i$.}\label{fig2}
\end{figure}

Each robot $i$ has two states: \textsc{Active} and \textsc{Passive} (see Fig. \ref{fig2}). Robot $i$ enters \textsc{Active} state when a task specification is active, and it switches to \textsc{Passive} state if and only if the task specification is completed. When robot $i$ is in \textsc{Passive} state, all the four modules are off and it will be viewed as a static obstacle. At the time instant that robot $i$ enters \textsc{Active} state, the motion planning module is activated and an initial (optimal) plan is synthesized (explained later). During online implementation, robot $i$ tries to satisfy its task specification safely by resolving conflicts with other robots. This is done by following some mode switching rules encoded into a Finite State Machine (FSM). Each FSM has the following three modes:

\begin{itemize}
  \item \textbf{Free}: Robot moves as planned. This is the normal mode, in which there is no conflict detected.
  \item \textbf{Busy}: Robot enters this mode when conflicts are detected.
  \item \textbf{Emerg}: Robot starts an emergency stop process.
\end{itemize}

In Fig. \ref{fig2}, the transitions between different modes of the FSM are also depicted. Initially, robot $i$ is in \textbf{Free} mode. Once conflict neighbors (will be defined later) are detected, robot $i$ switches to \textbf{Busy} mode and the motion planning module is activated to solve the conflicts, otherwise, robot $i$ stays in \textbf{Free} mode. When robot $i$ is in \textbf{Busy} mode, it switches back to \textbf{Free} mode if the motion planning module returns a feasible solution, otherwise (e.g., no feasible plan is found), robot $i$ switches to \textbf{Emerg} mode. Note that when robot $i$ switches to \textbf{Emerg} mode, it will come to a stop but with power-on. This means that robot $i$ will continue monitoring the environment and restart (switches back to \textbf{Free} mode) the task when it is possible.

\subsection{Initialization}

Denote by $t_0^i$ the task activation time of robot $i$. Then, robot $i$ enters \textsc{Active} state at $t_0^i$.
Once robot $i$ is \textsc{Active}, it first finds an optimal trajectory $\bm{\xi}_i$ (without the knowledge of other robots) such that the corresponding position trajectory $\textbf{p}_i\models \varphi_i$. The trajectory planning problem for a single robot can be solved by many existing methods, such as search/sampling based method \cite{Bhattacharya10,LaValle00}, automata-based method \cite{Quottrup04} and optimization-based method \cite{Howard10,Schulman13}. We note that the details of initial trajectory planning is not the focus of this paper. We refer to interested readers to corresponding literatures and the references therein.

\subsection{Decision making}

\subsubsection{Conflict detection}\label{section_cd}

Supposing that a cell decomposition is given over the workspace $\mathbb{X}$. The cell decomposition is a partition of $\mathbb{X}$
into finite disjoint convex regions $\Phi:=\{X_1, \ldots, X_{M_1}\}$ with $\mathbb{X}=\cup_{l=1}^{M_1} X_l$. Given a set $S\subset \mathbb{R}^{2}$, define the map $Q: \mathbb{R}^{2}\to 2^{\Phi}$ as
\begin{equation}\label{Qp}
Q(S):=\{X_l\in\Phi: X_l\cap S \neq \emptyset\},
\end{equation}
which returns the set of cells in $\Phi$ that intersect with $S$. The cell decomposition can be computed exactly or approximately using existing approaches (Chapters 4-5 \cite{Latombe12}). The choice depends on the particular models used for obstacles and other constraints.

The notation $\textbf{p}_i([t, \rightarrow))$ represents the position trajectory of robot $i$ from time $t$ onwards. Given a position trajectory $\textbf{p}_i([t_1, t_2])$ and a (set of) cell(s) ${\Phi}_l\subset \Phi$, the function $\Gamma: \mathcal{F}(\mathbb{R}_{\ge 0}, \mathbb{R}^2)\times 2^{\Phi} \to 2^{\mathbb{R}_{\ge 0}}$, defined as
\begin{equation}\label{Ti}
  \Gamma(\textbf{p}_i([t_1, t_2]), {\Phi}_l):=\{t\in [t_1, t_2]: \phi(\textbf{p}_i(t))\cap {\Phi}_l \neq \emptyset\},
\end{equation}
gives the time interval that robot $i$ occupies  ${\Phi}_l$. Let $t_c$ be the current time and $t_{i}^{fl}:=\min_{t>t_c}\{\textbf{p}_i(t)\notin \mathcal{B}(p_i(t_c), R)\}$ be the first time that robot $i$ leaves its sensing area $\mathcal{B}(p_i(t_c), R)$. Then, denote by $S_i(t_c):= \cup_{t\in [t_c, t_{i}^{fl}]}Q\big(\phi(\textbf{p}_i(t))\big)$
the set of cells traversed by robot $i$ within the time interval $[t_c, t_{i}^{fl}]$. Similarly, for each $j\in \mathcal{N}_i(t_c)$, let $t_{j}^{fl}$ be the first time that robot $j$ leaves its sensing area $\mathcal{B}(p_j(t_c), R)$ and $S_j(t_c)$ the set of cells traversed by robot $j$ within $[t_c, t_{j}^{fl}]$. Then, the \emph{conflict region} between robot $i$ and $j$ at time $t_c$ is defined as
\begin{equation}\label{Cij}
C_{i, j}(t_c)=S_i(t_c)\cap S_j(t_c).
\end{equation}

According to (\ref{Ti}), the time interval that robot $i(j)$ occupies the conflict cell $X_l\in C_{i, j}(t_c)$ is given by $\Gamma(\textbf{p}_i([t_c, t_{i}^{fl}]), X_l) \big(\Gamma(\textbf{p}_j([t_c, t_{j}^{fl}]), X_l)\big)$. Then, we have the following definition.

\begin{definition}\label{def2}
We say that there is a \emph{spatial-temporal conflict} between robot $i$ and $j$ at time $t$ if $\exists X_l\in C_{i, j}(t_c)$ such that $\Gamma(\textbf{p}_i([t_c, t_{i}^{fl}]), X_l) \cap \Gamma(\textbf{p}_j([t_c, t_{j}^{fl}]), X_l) \neq \emptyset$.
\end{definition}

Based on Definition \ref{def2}, define the set of conflict neighbors of robot $i$ at time $t_c$, denoted by $\tilde{\mathcal{N}}_i(t_c)$, as
\begin{equation*}
\begin{aligned}
  \tilde{\mathcal{N}}_i&(t_c):=\{j\in \mathcal{N}_i(t_c):\exists X_l\in C_{i, j}(t_c) \;\text{s.t.}\\ &\Gamma(\textbf{p}_i([t_c, t_{i}^{fl}]), X_l) \cap \Gamma(\textbf{p}_j([t_c, t_{j}^{fl}]), X_l) \neq \emptyset\}.
  \end{aligned}
\end{equation*}
%

Robot $i$ switches to \textbf{Busy} mode if and only if the set of conflict neighbors is non-empty (i.e., $\tilde{\mathcal{N}}_i(t_c)\neq \emptyset$). The conflict detection process is outlined in Algorithm 1.

\begin{algorithm}\label{algorithm1}
\caption{\emph{Conflict Detection}}
\begin{algorithmic}[1]
\Require $S_j(t_c), \Gamma(\textbf{p}_j([t_c, t_{j}^{fl}]), X_l),  \forall X_l\in S_j(t_c),  \forall j\in \mathcal{N}_i(t_c)\cup\{i\}$.
\Ensure $\tilde{\mathcal{N}}_i(t_c)$.
\State Initialize $\tilde{\mathcal{N}}_i(t_c)=\emptyset$.
\For {$j\in \mathcal{N}_i(t_c)$}
\If {$\exists X_l\in S_i(t_c)\cap S_j(t_c)$ s.t. $\Gamma(\textbf{p}_i([t_c, t_{i}^{fl}]), X_l) \cap \Gamma(\textbf{p}_j([t_c, t_{j}^{fl}]), X_l) \neq \emptyset$,}
\State $\tilde{\mathcal{N}}_i(t_c)=\tilde{\mathcal{N}}_i(t_c)\cup \{j\}$,
\EndIf
\EndFor
\If {$\tilde{\mathcal{N}}_i(t_c)\neq \emptyset$}
\State Robot $i$ switches to \textbf{Busy} mode.
\EndIf
\end{algorithmic}
\end{algorithm}
%

\subsubsection{Determine planning order}\label{section_pa}

Based on the neighboring relation, the graph $\mathcal{G}(t_c)=\{\mathcal {V}, \mathcal {E}(t_c)\}$ formed by the group of robots is naturally divided into one or multiple connected subgraphs, and the motion planning is conducted in parallel within each connected subgraph in a sequential manner. In order to do that, a planning order needs to be decided within each connected subgraph. In this work, we propose a simple rule to assign priorities between each pair of neighbors.

The number of neighbors of robot $i$ at time $t_c$ is given by $|\mathcal{N}_i(t_c)|$. Denote by $C_{i, \mathcal{N}_i(t_c)}(t_c):= \cup_{j\in \mathcal{N}_i(t_c)}\{C_{i, j}(t_c)\}$ the entire conflict region of robot $i$ at time $t_c$. Let
\begin{equation}\label{tmin}
  \underline{T}_i(t_c)=\min_{{X}_l\in C_{i,\mathcal{N}_i(t_c)}(t_c)}\left\{\min\{\mathcal{T}_i(X_l) \}\right\}
\end{equation}
be the earliest time that robot $i$ enters a conflict cell. Then, we have the following definition.

\begin{definition}\label{def3}
We say that robot $i$ has \emph{advantage} over robot $j$ at time $t_c$ if

1) $|\mathcal{N}_i(t_c)|>|\mathcal{N}_j(t_c)|$; OR

2) $|\mathcal{N}_i(t_c)|= |\mathcal{N}_j(t_c)|$ and $\underline{T}_i(t_c)<\underline{T}_j(t_c)$.
\end{definition}

Let $\mathcal{Y}_i(t_c)$ be the set of neighbors that have higher priority than robot $i$ at time $t_c$. The planning order determination process is outlined in Algorithm 2.



\begin{algorithm}\label{algorithm2}
\caption{\emph{Determine planning order}}
\begin{algorithmic}[1]
\Require $\mathcal{N}_i(t_c), P_i^0$ and $\underline{T}_j(t_c), P_j^0, j\in \mathcal{N}_i(t_c)$.
\Ensure $\mathcal{Y}_i(t_c)$.
\State Initialize $\mathcal{Y}_i(t_c)=\emptyset$.
\State Compute $\underline{T}_i(t_c)$ according to (\ref{tmin}),
\For {$j \in \mathcal{N}_i(t_c)$},
\If {\text{$j$ is in \textsc{Passive} state or \textbf{Emerg} mode}},
\State $\mathcal{Y}_i(t_c)=\mathcal{Y}_i(t_c)\cup j$,
\Else
\If {$j$ has advantage over $i$},
\State $\mathcal{Y}_i(t_c)=\mathcal{Y}_i(t_c)\cup j$,
\Else
\If {neither robot $i$ nor $j$ has advantage over the other and $P_j^0>P_i^0$},
\State $\mathcal{Y}_i(t_c)=\mathcal{Y}_i(t_c)\cup j$,
\EndIf
\EndIf
\EndIf
\EndFor
\end{algorithmic}
\end{algorithm}

\begin{proposition}[Deadlock-free]
The planning order assignment rule given in Algorithm 2 will result in no cycles, i.e., $\nexists \{q_m\}_1^{\hat k}, {\hat k}\ge 2$ such that $q_{\hat k}\in \mathcal{Y}_{q_1}$ and $q_{m-1}\in \mathcal{Y}_{q_m}, \forall m=2, \ldots, {\hat k}$.
\end{proposition}

\begin{remark}
The rationale behind our rule can be explained as follows. The total time required to complete the motion planning is given by $K\mathcal{O}(dt)$, where $\mathcal{O}(dt)$ represents the time complexity of one round of motion planning and $K$ represents the number of rounds (if multiple robots conduct motion planning in parallel, it is counted as one round), which is determined by the priority assignment rule being used (e.g., if fixed priority is used, the number of rounds is $K=N$). In our rule, we assign the robot with more neighbors the higher priority, and in this way, the minimal number of rounds can be achieved. Furthermore, if two conflict robots have the same number of neighbors, then the one that arrives earlier at the conflict region should have higher priority.
\end{remark}

\subsection{Motion planning}

The motion planning module consists of two submodules, i.e., fixed-path planning and trajectory planning. The fixed-path planning is activated first while the trajectory planning is activated if and only if the fixed-path planning returns no feasible solution.

\subsubsection{Fixed-path planning}

To define the fixed-path planning problem formally, the following notations are required. Denote by $\textbf{g}_i\subset \mathbb{R}^2$ the \emph{path} of robot $i$, which is a one dimensional manifold (curves in position space) that can be parameterized using the distance $s_i$ along the path, then $\textbf{g}_i(s_i)\in \mathbb{R}^2$. In this case, the path starts at $\textbf{g}_i(0)$, the tangent vector $\dot{\textbf{g}}_i(s_i)=\partial\textbf{g}_i/\partial s_i$ has unit length. The \emph{velocity profile} of robot $i$ is denoted by $s_i(t)$, which is a mapping from time to distance along the path.  Before starting to plan, robot $i$ needs to wait for the updated plan from the set of neighbors that have higher priority than robot $i$ (i.e., $j\in \mathcal{Y}_i(t_c)$) and consider them as moving obstacles. Denoted by $\textbf{p}_j^+([t_c, \rightarrow))$ the updated position trajectory of robot $j$ and let $t_{j}^{fl+}$ be the first time that robot $j$ leaves its sensing area  $\mathcal{B}(p_j(t_c), R)$ according to $\textbf{p}_j^+([t_c, \rightarrow))$. Then, one can define $ S_j^+(t_c)$ as the set of cells traversed by robot $j$ within $[t_c, t_{j}^{fl+}]$.

\begin{definition}\label{def1}
We say there is a \emph{spatial conflict} between robot $i$ and $j, j\in \mathcal{Y}_i(t_c)$ before (after) the fixed-path planning if $S_i(t_c)\cap S_j^+(t_c)\neq \emptyset$ ($S_i^+(t_c)\cap S_j^+(t_c)\neq \emptyset$).
\end{definition}

Due to the fixed-path property, one can conclude that $S_i(t_c)\cap S_j^+(t_c)=\emptyset \Rightarrow S_i^+(t_c)\cap S_j^+(t_c)=\emptyset$. Therefore, in fixed-path planning, only the higher priority neighbors that have spatial conflict with robot $i$ before the fixed-path planning need to be considered. Based on this observation, we define $C_{i,j}^+(t_c):=S_i(t_c)\cap S_j^+(t_c)$ as the (updated) set of conflict cells between robot $i$ and $j$ at time $t_c$ and let $\tilde{\mathcal{Y}}_i(t_c):=\{j\in \mathcal{Y}_i(t_c): C_{i, j}^+(t_c)\neq \emptyset\}$ be the set of higher priority robots that have spatial conflict with robot $i$. Denote by $\textbf{g}_i^{t_c}:=\textbf{p}_i([t_c, \rightarrow))$ the path of robot $i$ at time $t_c$. Then, the fixed-path planning problem (FPPP) is formulated as follows:
\begin{subequations}\label{optim1}
\begin{eqnarray}
&&\hspace{-0cm}\min  \quad J_i(\xi_i, u_i)\\
&&\hspace{-1.5cm}\text{subject to}\nonumber \\
&&\hspace{-1cm}\dot{\textbf{g}}_i^{t_c}(s_i)\dot s_i\in \Pi(\textbf{g}_i^{t_c}),\label{optim1-a}\\
&&\hspace{-1cm} \begin{aligned} \textbf{g}_i^{t_c}(s_i(t))\notin &X_l,\; t\in \Gamma(\textbf{p}_j^+([t_c, t_{j}^{fl+}]), X_l),\\
 &\forall j\in \tilde{\mathcal{Y}}_i(t_c), \forall X_l\in C_{i,j}^+(t_c), \end{aligned} \label{optim1-c}
\end{eqnarray}
\end{subequations}
where $s_i$ is the speed profile that needs to be optimized and $\Pi(\textbf{g}_i^{t_c})$ is defined as
\begin{equation*}
\begin{aligned}
  \Pi(\textbf{g}_i^{t_c})&:=\{\dot{\textbf{g}}_i^{t_c}(s_i)\dot s_i: \exists v_i\in \mathbb{V}_i, \omega_i\in \mathbb{W}_i, u_i\in \mathbb{U}_i, s.t., \\
  &\dot{\textbf{g}}_i^{t_c}(s_i)\dot s_i=(\cos(\theta_i), \sin(\theta_i))v_i, \dot\theta_i=\omega_i, (\dot v_i, \dot\omega_i)=u_i\}.
  \end{aligned}
\end{equation*}

\subsubsection{Trajectory planning}

If fixed-path planning returns no feasible solution, then it is necessary to replan the trajectory (path and velocity profile). The trajectory planning problem (TPP) can be formulated as follows:
\begin{subequations}\label{optim}
\begin{eqnarray}
&&\hspace{-0cm}\min  \quad J_i(\xi_i, u_i),\\
&&\hspace{-1.1cm}\text{subject to}\nonumber \\
&&\hspace{-1cm} (\ref{c3}),\\
&&\hspace{-1cm} \begin{aligned} \verb"proj"_{p_i}&(\xi_i(t))\notin X_l,\; t\in \Gamma(\textbf{p}_j^+([t_c, t_{j}^{fl+}]), X_l),\\
& \forall j\in \mathcal{Y}_i(t_c), \forall X_l \in S_j^{+}(t_c) \cap Q(\mathcal{B}(p_i(t_c), R)). \end{aligned} \label{optim-b}
\end{eqnarray}
\end{subequations}

If both FPPP (\ref{optim1}) and TPP (\ref{optim}) return no feasible solution, robot $i$ switches to \textbf{Emerg} mode. In this mode, robot $i$ will continue monitoring the environment, and once the
TPP (\ref{optim}) becomes feasible, it will switch back to \textbf{Free} mode. The motion planning process is outlined in Algorithm 3.

\begin{remark}
Various existing optimization toolboxes, e.g., IPOPT \cite{ipopt}, ICLOCS2 \cite{iclocs}, and algorithms, e.g., the configuration space-time search \cite{Parsons90} and the Hamilton-Jacobian reachability-based motion planning \cite{Mo18} can be utilized to solve (\ref{optim1}) and (\ref{optim}). We note that, in general (no matter which method is used), the computational complexity of TPP (\ref{optim}) is much higher than that of FPPP (\ref{optim1}).
\end{remark}

\begin{algorithm}\label{algorithm3}
\caption{\emph{Motion Planning}}
\begin{algorithmic}[1]
\Require $S_j^+(t_c), \Gamma(\textbf{p}_j^+([t_c, t_{j}^{fl+}]), X_l), \forall j\in \mathcal{Y}_i(t_c)\cup \{i\}, \forall X_l\in S_j^+(t_c)$.
\Ensure $\textbf{p}_i^+([t_c, \rightarrow))$.
\State Compute $\tilde{\mathcal{Y}}_i(t_c)$ and solve the FPPP (\ref{optim1}),
\If {Solution obtained (denoted by $s_i^*$),}
\State $\textbf{p}_i^+([t_c, \rightarrow))=\textbf{g}_i^{t_c}(s_i^*)$,
\State Robot $i$ switches to \textbf{Free} mode,
\Else
\State Solve the TPP (\ref{optim}),
\If {Solution obtained (denoted by ${\bm{\xi}}_i^*$)},
\State $\textbf{p}_i^+([t_c, \rightarrow))=\verb"proj"_{p_i}({\bm{\xi}}_i^*)$,
\State Robot $i$ switches to \textbf{Free} mode,
\Else
\State Robot $i$ switches to \textbf{Emerg} mode,
\If {The TPP (\ref{optim}) is feasible,}
\State Robot $i$ switches to \textbf{Free} mode,
\EndIf
\EndIf
\EndIf
\end{algorithmic}
\end{algorithm}

\begin{remark}
Due to the distributed fashion of the solution and the locally available information, the proposed MRMC strategy is totally scalable in the sense that the computational complexity of the solution is not increasing with the number of robots. In addition, it is straightforward to extend the work to MRSs scenarios where moving obstacles are presented.
\end{remark}

\begin{remark}
We assume that the deceleration (i.e., negative force input) that each robot can take when switching to \textbf{Emerg} mode is unbounded. This guarantees that no inter-robot collision will occur during the emergency stop process since in the worst case, the robot can stop immediately. The problem of safety guarantees under bounded deceleration in  \textbf{Emerg} mode will be studied in future work.
\end{remark}


\section{Simulation}

\begin{figure}[h!]
\centering
\includegraphics[height=6cm,width=8cm]{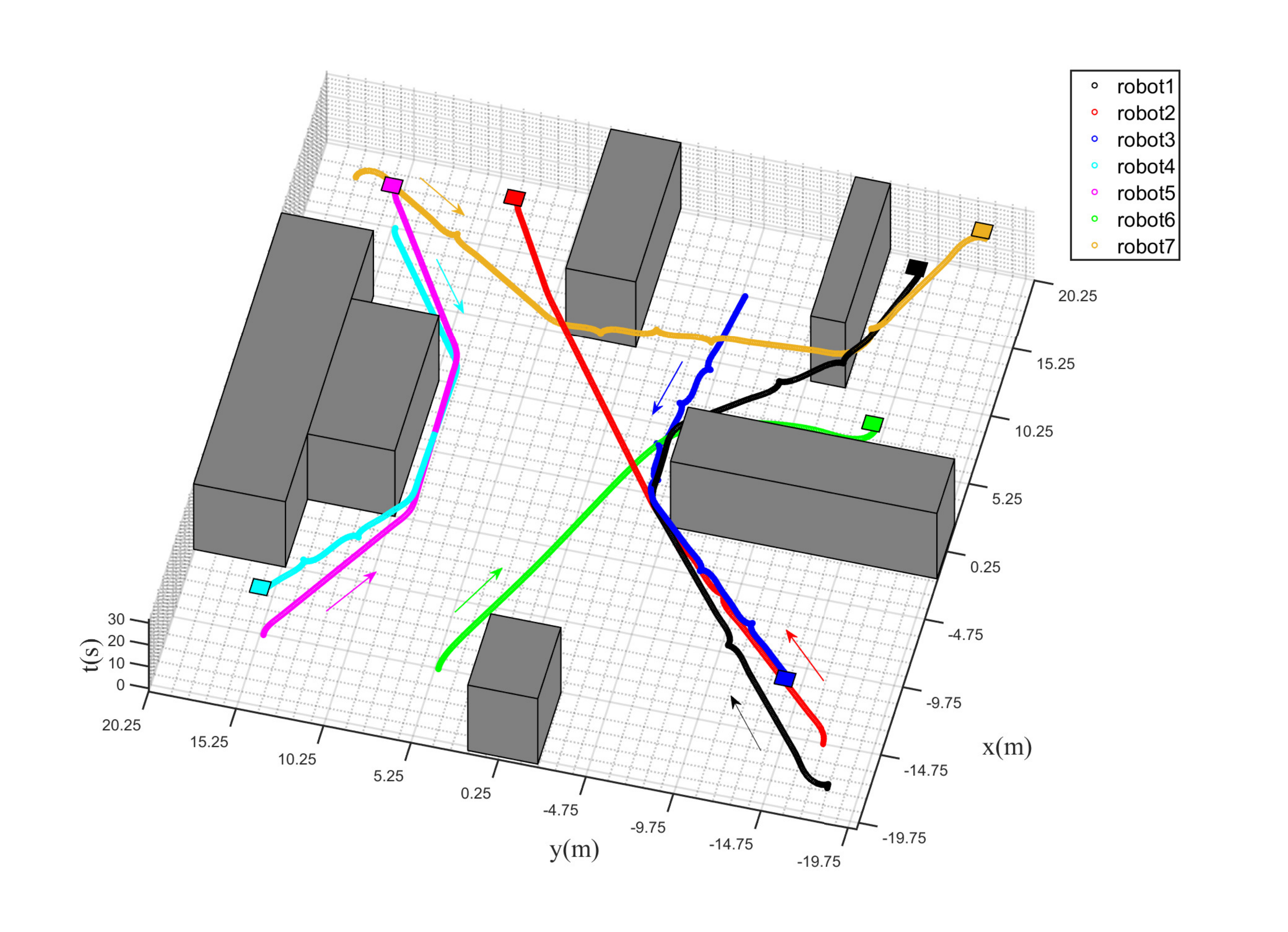}
\caption{The initial optimal trajectories of each robot. }\label{fig5}
\end{figure}

We illustrate the results of the paper on a MRS consisting of $N=7$ robots. The velocity and input constraints for each robot are given by $|v_i|\le 2 m/s, |\omega_i|\le 115 rad/s, |F_i|\le 2 m/s^2, |\tau_i|\le 115 rad/s^2$ and the sensing radius is $R=5m$. The common workspace for the group of robots is depicted in Fig. \ref{fig5} ($xy$ axis), where the gray areas represent the obstacles and a grid representation with grid size $0.5m$ is implemented as cell decomposition. For each robot, the task specification is given by $\varphi_i:=\square \neg \mathbb{O} \wedge \lozenge \square X_i^f, \forall i$, where $X_i^f$ (marked as colored region in Fig. \ref{fig5}) represents the target set for robot $i$, while $\neg$, $\square$ and $\lozenge$ are respectively ``negation", ``always" and ``eventually" operators in a linear temporal logic formula.

\begin{figure}[h!]
\centering
\includegraphics[height=6cm,width=8cm]{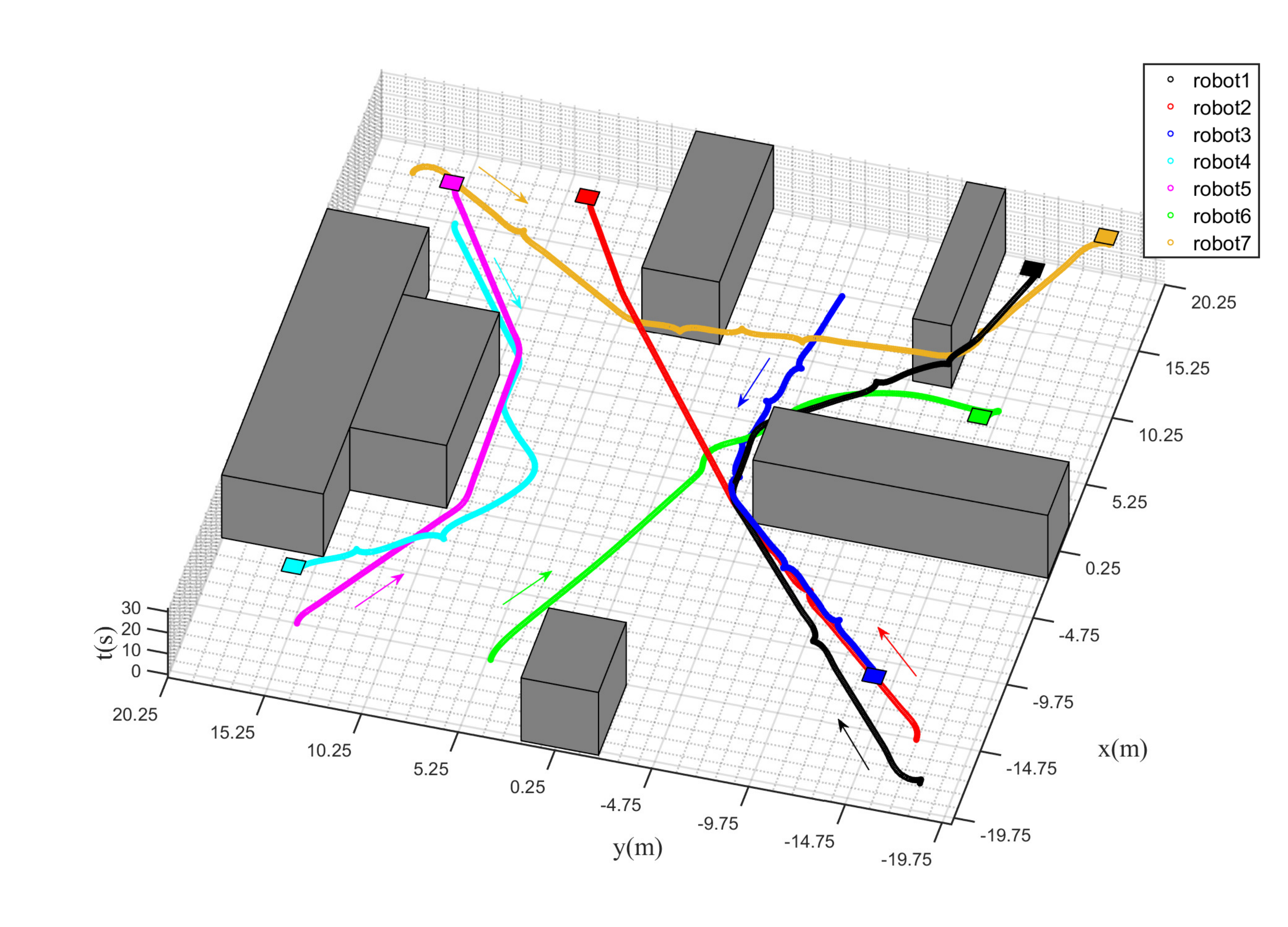}
\caption{The real-time trajectories of each robot. }\label{fig6}
\end{figure}

The initial optimal trajectories for each robot are depicted in Fig. \ref{fig5}, where the colored arrows show the moving direction of each robot. It can be seen that conflicts (i.e., inter-robot collisions) occur between robot pairs $(1, 3), (1, 7), (4, 5)$ and $(3, 6)$. During online implementation, conflicts are detected by each robot and replanning is conducted by robots 1, 4 and 6 at time instants $19.7s, 7.5s$ and $8.4s$, respectively. In the motion planning module, the ICLOCS2 \cite{iclocs} toolbox is implemented to solve the FPPP (\ref{optim1}) and the TPP (\ref{optim}). The real-time moving trajectory for each robot is shown in Fig. \ref{fig6}, where all the conflicts are resolved.


\begin{figure}[h!]
\centering
\includegraphics[height=4cm,width=9cm]{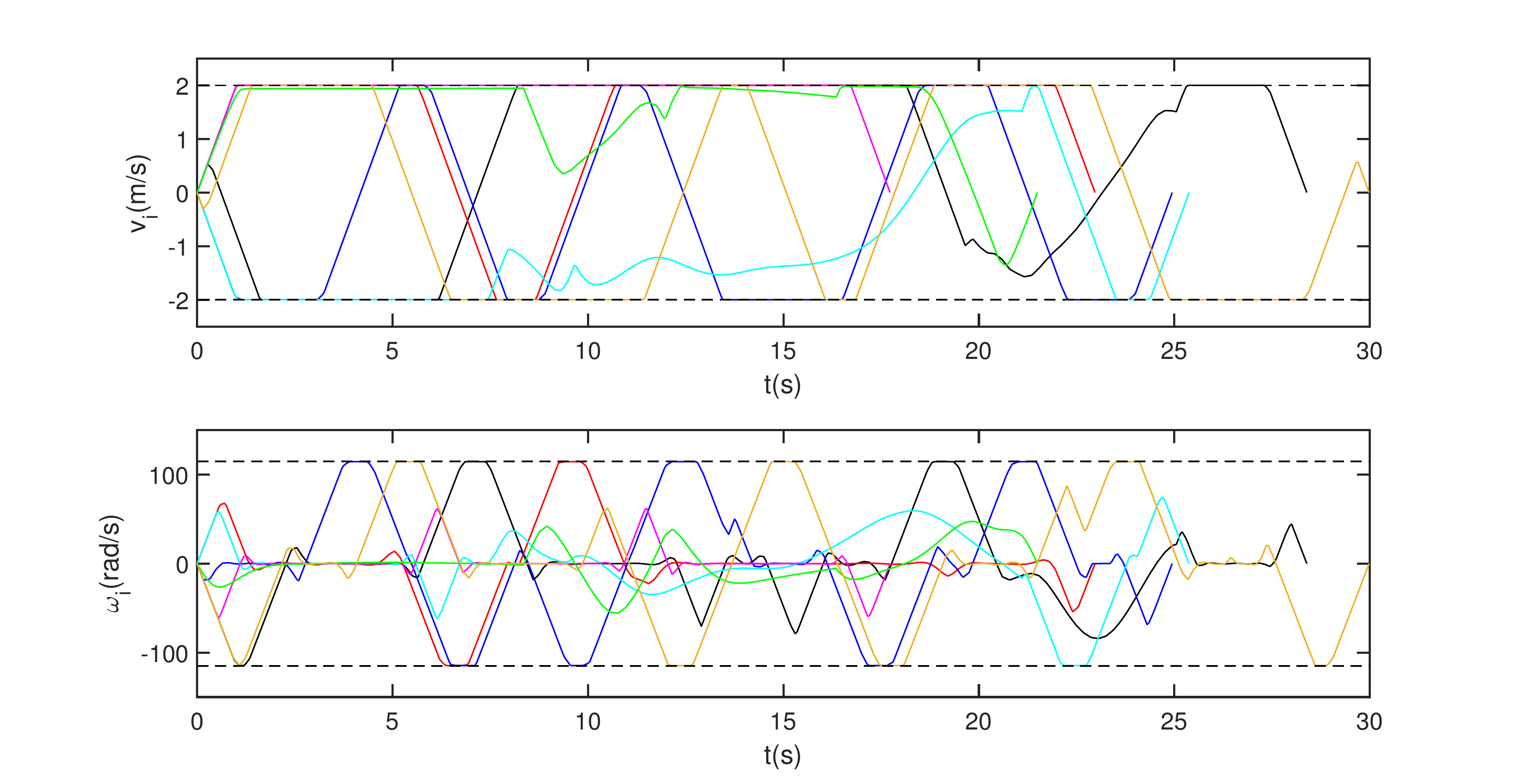}
\caption{The real-time evolution of $v_i$ and $\omega_i$. }\label{fig7}
\end{figure}

\begin{figure}[h!]
\centering
\includegraphics[height=4cm,width=9cm]{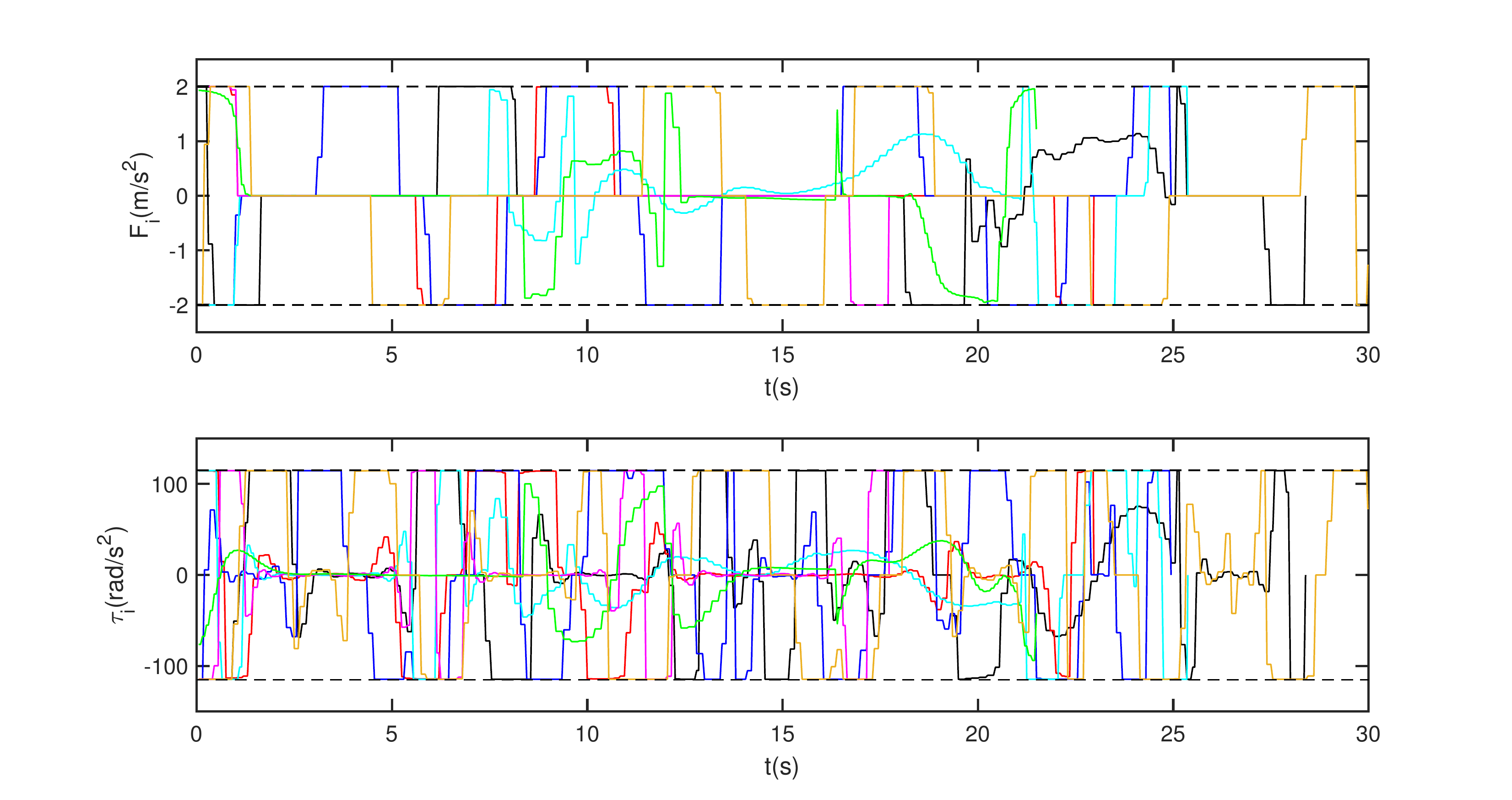}
\caption{The real-time evolution of $F_i$ and $\tau_i$.} \label{fig8}
\end{figure}

The real-time evolution of velocities $(v_i, \omega_i)$ and inputs $(F_i, \tau_i)$ for each robot are given in Fig. \ref{fig7} and Fig. \ref{fig8}, respectively. One can see that the tangential and angular velocity constraints and the force and torque input constraints are satisfied by all robots at all times. All the simulations were run in Matlab 2018b on a DELL laptop of 2.6GHz using Intel Core i7.

\section{Conclusion}

In this paper, the online MRMC problem is considered. Under the assumptions that each robot has only local view and local information, and subject to both velocity and input constraints, a fully distributed motion coordination strategy was proposed for steering individual robots in a common workspace, where each robot is assigned independent task specifications. It was shown that the proposed strategy can guarantee collision-free motion of each robot. A next step is to perform real-world experiments.

\section*{Acknowledgement}

The authors would like to thank Yulong Gao for valuable discussions.


\end{document}